# Enhanced Large Language Models for Effective Screening of Depression and Anxiety

June M. Liu [a,b +], Mengxia Gao [a,b +], Sahand Sabour [c], Zhuang Chen [c], Minlie Huang [c #], Tatia M.C. Lee [a,b #]

[a] State Key Laboratory of Brain and Cognitive Sciences, The University of Hong Kong, Hong Kong, China.
[b] Laboratory of Neuropsychology and Human Neuroscience, The University of Hong Kong, Hong Kong, China.
[c] The CoAI group, DCST; Institute for Artificial Intelligence; State Key Lab of Intelligent Technology and Systems; Beijing National Research Center for Information Science and Technology; Tsinghua University, Beijing, China.

[+] These authors contributed equally to this article

# **Correspondence:**
Tatia M.C. Lee, Ph.D.
Address: Room 656, Laboratory of Neuropsychology & Human Neuroscience, The Jockey Club Tower, The University of Hong Kong, Pokfulam Road, Hong Kong.
E-mail address: tmclee@hku.hk

Minlie Huang, Ph.D.
Address: Room 4-504, FIT Building, Dept. of Computer Science, Tsinghua University, Beijing 100084, China
China. E-mail: aihuang@tsinghua.edu.cn

**Abstract**

Depressive and anxiety disorders are widespread, necessitating timely identification and management. Recent advances in Large Language Models (LLMs) offer potential solutions, yet high costs and ethical concerns about training data remain challenges. This paper introduces a pipeline for synthesizing clinical interviews, resulting in 1,157 interactive dialogues (PsyInterview), and presents EmoScan, an LLM-based emotional disorder screening system. EmoScan distinguishes between coarse (e.g., anxiety or depressive disorders) and fine disorders (e.g., major depressive disorders) and conducts high-quality interviews. Evaluations showed that EmoScan exceeded the performance of base models and other LLMs like GPT-4 in screening emotional disorders (F1-score=0.7467). It also delivers superior explanations (BERTScore=0.9408) and demonstrates robust generalizability (F1-score of 0.67 on an external dataset). Furthermore, EmoScan outperforms baselines in interviewing skills, as validated by automated ratings and human evaluations. This work highlights the importance of scalable data-generative pipelines for developing effective mental health LLM tools.

# Introduction

Emotional disorders are among the most prevalent mental health conditions worldwide (Bullis et al., 2019; Institute of Health Metrics and Evaluation, 2019), leading to at least $6.5 trillion healthcare-related costs globally (Konnopka & König, 2022). Among these conditions, anxiety and depressive disorders are particularly common (Finning et al., 2017; Zvolensky et al., 2014; Watson et al., 2008), with a high prevalence rate of approximately 4.8% and 3.2% respectively, in the general population (Santomauro et al., 2021). While timely screening of emotional disorders, followed by an appropriate treatment, is crucial for individuals' well-being (Sau & Bhakta, 2019), this process can be time-consuming and labor-intensive, often requiring comprehensive interviews, collection of history and background information, and significant manpower from healthcare professionals (Wright et al., 2017; American Psychiatric Association, 2015).

To address this challenge, we proposed to utilize Large Language Models (LLMs) to facilitate the screening process. LLMs are deep learning systems trained on massive collections of text to predict words in a sequence (Blank, 2023). There has been a growing interest in applying LLMs to solve mental health-related problems (Ke et al., 2024; Omar et al., 2024; King et al., 2023; Stade et al., 2023). Existing studies have explored the potential of using textual information to identify psychiatric symptoms and significant life events (Chen et al., 2024a), detect depression based on post histories on social media (Lan et al., 2024), and provide explainable screening for emotional disorders (Wang et al., 2024; Xu et al., 2024). However, most existing studies mainly focus on using data from online platforms (i.e., social media) for screening emotional disorders. While these studies show the potential of utilizing LLMs for large-scale screenings, they may not be applicable in clinical contexts due to the lack of detailed backgrounds, symptoms, and diagnoses of individuals from real clinical cases.

Brief clinical interviews could be a more reliable method for screening compared to using data from online platforms since they provide an opportunity to gather more relevant information (Nie et al., 2024). Previous studies have leveraged semi-structured clinical interviews or questions from scales to obtain clients' information, which were further used for identifying emotional disorders (Rosenman et al., 2024; Chen et al., 2024b). Recognizing the effectiveness of these methods, efforts have been made to develop LLMs incorporating real-life clinical interviews and expert diagnoses (Tu et al., 2024). Although the study of utilizing clinical interviews for developing LLMs has shown potential, the process of conducting interviews for data collection is highly time-consuming and expensive, making it less feasible for training large-scale models. Moreover, the sensitive nature of human studies and stringent privacy regulations often restrict the direct use of real clinical data (Adarmouch et al., 2020). Consequently, there is a critical need for a scalable approach to synthesize data that can emulate real-world scenarios without violating participants' confidentiality. However, the challenge remains in generating sufficiently large and diverse interviewing data that accurately capture nuanced information from clients with emotional disorders.

In light of the aforementioned challenges, we proposed a novel data-generative pipeline that synthesizes clinical interviews on emotional disorders, thereby facilitating the development of clinical LLMs (Fig. 1a). This pipeline was built following comprehensive guidelines of psychiatric interviews. Utilizing this pipeline, we generated PsyInterview, a dataset derived from casebooks and clinical notes, which comprises multi-turn interviews, corresponding screenings, and accompanying explanations. By automating the generation of clinical interview data, our approach could address the scarcity and sensitivity of real-world clinical data, enabling the training of LLMs on a larger and more diverse set of scenarios.

Utilizing the PsyInterview dataset, we developed an LLM agent specifically designed for screening emotional disorders from clinical interviews (Fig. 1b). It can not only distinguish between coarse disorders such as anxiety disorders and depressive disorders but also identifies fine-grained conditions such as Major Depressive Disorder and Generalized Anxiety Disorder. Importantly, the agent can provide explanations for its screening results, enhancing transparency and trustworthiness. Moreover, the PsyInterview can also be leveraged to train an interviewing agent, which can serve as a virtual assistant to reduce psychiatrists or clinical psychologists' burden by automating the initial interviewing process (Fig. 1c). To make sure about the interviewing quality, we assessed its performance using both automatic and human experts' evaluation. We combined the screening and interviewing agents as an LLM-based system, EmoScan, for emotional disorders' screening and interview-assisting. Our experiments demonstrated the high quality of PsyInterview and the efficacy of EmoScan in performing screenings and conducting interviews. Overall, this study highlights the capabilities of our novel generative pipeline in synthesizing interviewing data and the potential of EmoScan for screening and interviewing individuals with emotional disorders. An overview of the study is presented in Fig. 1.

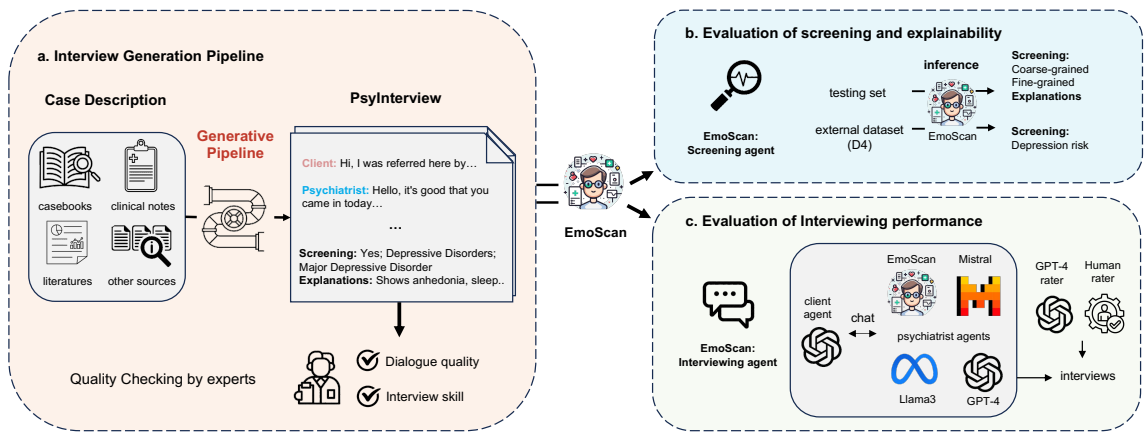

**Fig. 1 Overview of the Study. a.** Our proposed generative pipeline, which transforms various formats of case descriptions/information to clinical interview. We recruited licensed psychiatrists and clinical psychologists to evaluate the quality of the generated interview. Then we created a system, EmoScan, consisting of two agents designed to screen for emotional disorders while providing relevant explanations, and conducting brief clinical interviews, respectively. **b.** The screening agent can screen for emotional disorders based on the conversational history. We evaluated its screening performance on the testing dataset and also its generalization on an external dataset (D4). **c.** The interviewing agent is designed to communicate effectively with users. We evaluated the interviewing performance of the agent conducting pairwise comparisons with other LLMs. Briefly, we instructed GPT-4 to act as a client and chat with the studied LLMs (EmoScan, Mistral, Llama3, and GPT-4). Subsequently, another GPT-4 rater and human experts separately reviewed the conversation history to assess the interviewing skills of these LLMs.

## Methods

### Data Preparation

To acquire high-quality data for training effective LLMs in clinical settings, we initially developed a four-stage data generative pipeline that transforms various forms of case descriptions or information into refined psychiatrist-client dialogues. Subsequently, we gathered 1,157 cases involving emotional and other psychiatric disorders and converted them into interactive interviews utilizing our data-generative pipeline. To ensure the data quality of the generated interview data, we recruited three clinical psychologists to evaluate the generated dialogues. Finally, we presented the dataset PsyInterview. Details of the data preparation are shown below.

#### *Data Generative Pipeline*

We have developed a data-generative pipeline designed to transform varying formats of case descriptions or information with reliable labels into polished psychiatrist-client dialogues. These refined dialogues can later be used to train models or conduct evaluations.

The pipeline comprises four main stages (Fig. 2.). The first step requires gathering detailed client information or descriptions. This data can be extracted from casebooks and clinical notes describing the client's experiences or stories, and also from scientific literature and databases containing dialogue resources. Once the raw data is compiled, the next step involves extracting key components such as the client's complaints, medical history, history of drug or alcohol abuse, etc. To achieve this, we adapted a standardized template (Prendergast, 2018) conventionally employed for psychiatric evaluation. This approach ensures a thorough extraction of the client's complete background. The details of the template can be found in Appendix 1. The third step involves converting the extracted data into a raw conversation following a topic flow based on Morrison's (2016) guidebook for psychiatric interviews. Briefly, the psychiatrist will first ask the client's identification and complaints then collect the medical and psychiatric histories, family history, and finally personal and social history. The final stage of the pipeline focuses on polishing the raw conversation; for example, removing sensitive personal information (e.g., name and location) and deleting duplicate content. The polishing rules were modified from those in Wang et al.'s (2023) study on synthetic clinical interviewing conversations. Prompts for each of these steps are comprehensively provided in Appendix 2.

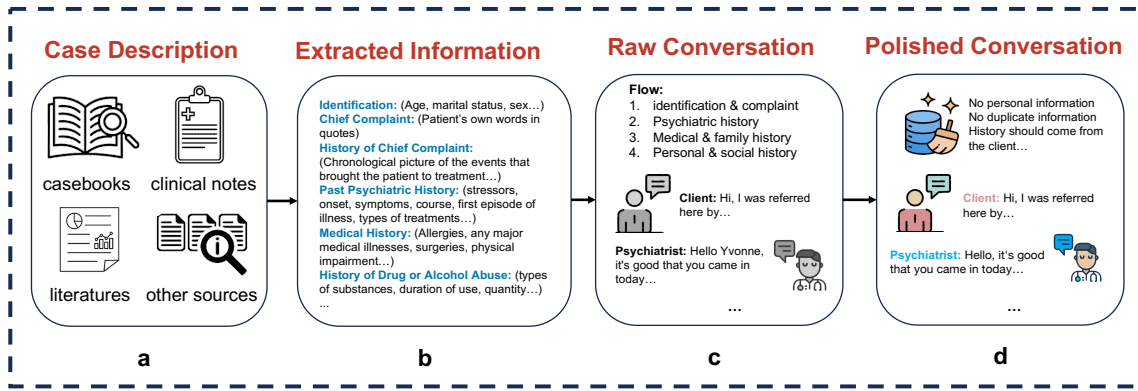

**Fig. 2 The Interview Generative Pipeline. a.** Collect case descriptions from clinical casebooks, clinical notes, scientific literatures, and other related sources. **b.** Extract information from the case description following a screening template. **c.** Generate raw interviews from the extracted information. The conversations should follow an interviewing flow. **d.** Polish the generated conversations to remove private information and prevent unexpected information leakage.

*Data Source*

  To train a model capable of identifying emotional disorders in a diverse population, we collected cases encompassing emotional disorders, other mental disorders of the Diagnostic and Statistical Manual of Mental Disorders, Fifth Edition (DSM-5) excluding depressive and anxiety disorders (e.g., Schizophrenia Spectrum and Other Psychotic Disorders), and healthy controls. Case descriptions were collected from a wide array of sources, including clinical casebooks, research papers, and an open-source dataset (Cheng et al., 2023), which collectively provided us with a diverse range of 1,157 cases. Out of the 1,157 cases, there were 144 cases with emotional disorders, including depressive and anxiety disorders, and 269 cases with other mental disorders such as schizophrenia spectrum and other psychotic disorders. Given that mental disorders often exhibit shared clinical symptoms, including cases of other mental disorders could enhance the LLMs' ability to effectively characterize individuals with emotional disorders. By incorporating a broader range of cases, the LLMs can learn how to distinguish between different mental disorders. The screening results and explanations for these cases adhere to the diagnostic criteria outlined in the DSM-5. A detailed list of casebooks' and research papers' sources were provided in Appendix 3.

  The remaining 744 healthy control cases were sourced from the PESConv dataset (Cheng et al., 2023). The PESConv dataset was adapted from the ESConv dataset (Liu et al., 2021), which was created by recruiting crowd-workers and instructing them to engage in conversations while acting as help-seekers and supporters, with the goal of producing more natural dialogues that closely mimic real-life situations. ESConv was primarily developed for providing emotional support for non-clinical individuals facing common emotional concerns. Based on ESConv, PESConv extracted the cases' persona, which provided additional important information (i.e., demographic information, social status, personality, etc.) about the cases' identification. These personas enrich the clients' profile, thereby enhancing the informativeness of the responses within emotional support conversations. In this case, we adapted the extracted persona and conversations to serve as case information for healthy controls.

We then used the interview generative pipeline to transform the 1,157 case descriptions to corresponding interactive interviews. The sample distribution of the cases can be found in Fig. 3a. On average, each conversation consists of 14 utterances from either a psychiatrist or a client, with each utterance having 24 words. The detailed statistics of the synthetic interviews are shown in Fig. 3b.

*Data Quality-check*

To ensure the authenticity of the generated conversations in clinical settings, we recruited three experts to evaluate the quality of the generated dialogues. All three experts were clinical psychologists registered at the Hong Kong Psychological Society (HKPS). We randomly selected 50 cases from the training dataset, with approximately one-third of these cases representing depressive disorders, anxiety disorders and healthy control each. Each case was rated by two experts based on 1) information alignment between case description/information, dialogue, and explanations; 2) naturalness and consistency of the dialogues; and 3) logicality and compliance of explanations. The experts were asked to rate each item on a 5-point Likert scale (1 = very misaligned / not natural at all / etc., 5 = very aligned / very natural / etc.).

To further assess the reliability of our data, we adopted an interview skill assessment developed by Morrison (2016). Since the original assessment was designed for extended interviews and our project focused on relatively shorter dialogues (8-rounds), we selected dimensions aligned with our context, including history, ending the interview, establishing rapport, and use of interviewing techniques. History consists of the psychiatrist's inquiry about medical history, family history of mental disorder, history of present illness, and personal and social history. Ending the interview involves giving a warning that the interview is concluding and expressing interest and appreciation at the end. Establishing rapport focuses on building relationships with the client while the use of interview techniques dimension refers to the psychiatrist's approach in gathering client information. The history and ending the interviewing dimensions are particularly important for our interviewing agent, as obtaining an accurate history is crucial for achieving a reliable screening result of the emotional disorder (Jansson & Nordgaard, 2016), and employing proper ending tactics ensures the conversation remains focused and does not become endless. Meanwhile, the two supplementary dimensions, establishing rapport and use of interview techniques, more suited for longer interviews, ensure that the interviewing style of our generated corpus is acceptable. We collected the responses from the three experts and computed an average score for each of the dimensions. If the average score of a particular dimension exceeds 50% of the maximum, it indicates that the responses of that dimension are generally positive and acceptable. The results showed that our data received positive responses across all dimensions, as shown in Fig. 3c.

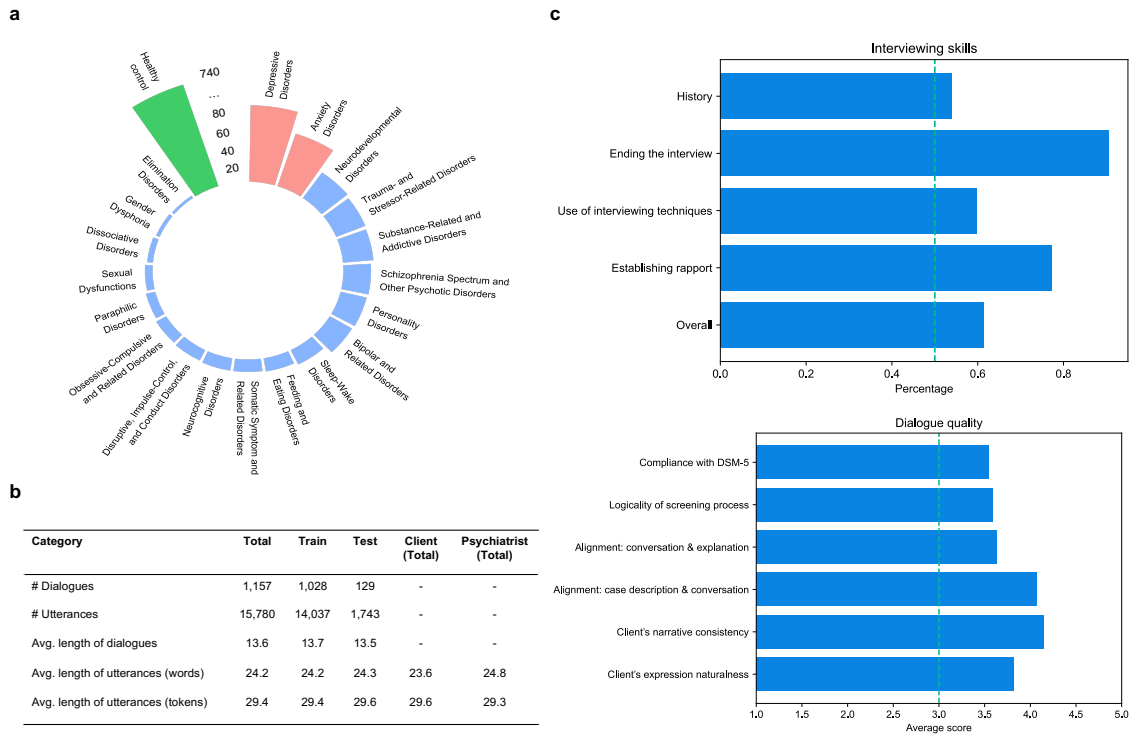

**Fig. 3 Data Distribution and Quality Evaluation. a.** Disorder distribution of the PsyInterview. The green bar represents healthy controls, the two pink bars correspond to emotional disorders, and the remaining blue bars indicate other disorders. **b.** Statistics of the PsyInterview, including the number of dialogues, average dialogue length, and average utterance length in terms of both word and token. **c.** Data quality-check results, comprising average scores for dialogue quality and interviewing skill. The green line denotes the threshold line.

**Model Training and Evaluation**

Using the generated interview data PsyInterview, we trained a system called EmoScan and evaluated its effectiveness in screening for emotional disorders. On the other hand, we assessed its performance in conducting interviewing tasks with GPT-4 acting as the client. In this section, we outline the training procedures for EmoScan's screening and interviewing agents, as well as the evaluation methods used to compare EmoScan's performance with baselines.

*Training*

We developed a system that consists of a screening and an interviewing agent. The screening agent was a fine-tuned Mistral-7B (Jiang et al., 2023) trained by conversational history and screening outputs (i.e., a combination of screening results and explanations), while the interviewing agent was trained by the same base-model with the multi-turn conversational data synthesized by the generative pipeline, as Mistral-7B and related models excel in diverse benchmarks (Jiang et al., 2023) and clinical tasks (Cong et al., 2024; Longwell et al., 2024). The screening agent will provide a result with an explanation that describes why the client gets the result according to the DSM-5 based on the conversational history, and the interviewing agent is capable to conduct psychiatric interviews with clients. Training details can be found in Appendix 4.

*Evaluation*

**Baselines.** We compared EmoScan with recent widely used LLMs, OpenAI's GPT-4 (gpt-4-0613) (OpenAI, 2023), Llama 3 (Meta-Llama-3-70B), and Mistral-7B (Jiang et al., 2023). These three LLMs served as baselines in screening/explainability and interviewing evaluation.

***Research Question 1: Does the screening agent have the ability to do screening and provide explanations?*** We compared the screening performance of EmoScan and baselines for both coarse- and fine-grained classification of emotional disorders. At the coarse-grained level, we assessed the LLMs' ability to identify individuals with depressive disorders or anxiety disorders from the healthy controls. At the fine-grained level, we evaluated the LLMs' capability to identify specific emotional disorders (e.g., Major Depressive Disorder, Generalized Anxiety Disorder) among positive cases. The classification performance was evaluated using weighted F1 (Schultebraucks et al., 2022), which indicates the harmonic mean of precision and recall depending on each class's sample size. To determine if there were significant differences in classification performance among the LLMs, we ran each model on the test set three times and then applied independent two-sample t-tests to assess the statistical significance of the performance differences between EmoScan and the baselines. Additionally, we used BERTScore (Zhang et al., 2019), ROUGE (Lin, 2004), and BLEU (Papineni, 2002) to measure the quality of explanations generated by the four LLMs, comparing their similarity to the ground truth explanations.

To explore the potential of baselines, we further employed few-shot (Brown et al., 2020) and chain-of-thought (CoT) (Wei et al., 2022) prompting techniques. Few-shot prompting provides LLMs with a few task examples during inference, which can improve LLMs' performance without changing their weights. In our study, we randomly selected four cases from the training data to serve as few-shot samples. The samples included diverse cases, encompassing individuals with depressive disorder, anxiety disorder, other disorders, and healthy controls. CoT prompting offers intermediate reasoning steps for LLMs, enhancing their performance on complex reasoning tasks. Our study adapted an emotional disorders' screening guideline from a comprehensive mental health disorder diagnosis handbook (Morrison, 2023) as the CoT prompt. The detailed CoT prompt can be found in Appendix 5.

To test the generalizability of EmoScan, we compared our system with the base model on an external out-of-domain dataset, D4 (Yao et al., 2022). This dataset was developed to screen for depression in simulated conversations between two crowdsource workers. Since this dataset was originally in Chinese, we first translated the dataset to English using the Youdao API (https://fanyi.youdao.com/openapi/)[1], and then compared the screening performance of our model and the base model Mistral-7B.

---
[1] We have conducted translation checking on the testing dataset.

***RQ2: Does the interviewing agent have acceptable interviewing performance?***
Obtaining essential key information about the client during an interview is crucial for achieving accurate screening results. To evaluate the interviewing performance of different models, we applied two fundamental dimensions in the interviewing assessment rated by experts: history and ending the interview. We compared EmoScan with baselines on these two essential dimensions.

To simulate the interviewing process between clients and psychologists, we first built a patient simulator using Autogen (Wu et al., 2023) to interact with all four LLMs (i.e., EmoScan, GPT-4, Llama 3, Mistral-7B). During this process, we instructed GPT-4 to act as a client, responding to the psychiatrist's questions, while providing it with the relevant patient information. Simultaneously, we assigned one of the four LLMs to act as the psychiatrist, responsible for asking questions. The interviewing dialogues were recorded for subsequent evaluation.

Previous studies have applied GPT-4 to evaluate conversational responses, demonstrating a strong correlation with human raters (Liao et al., 2024; Hackl, 2023). Therefore, in this study, we also used a separate GPT-4 agent to act as a judge, assessing the interviewing performance through a comparative analysis of the interviewing dialogues generated by our model versus those produced by the other three models. For each evaluation, raters were presented with two conversations generated by EmoScan–interviewing agent and the other model, in which the LLMs acted as psychiatrists and talked with the same simulated client. Models' names were masked, and the two conversations' order was randomized each time to avoid bias. After that, the rater voted for one of them on the two dimensions. Finally, we calculated the winning rate of the four models. To ensure the reliability of GPT-4's rating, we randomly selected 90 conversation pairs and recruited six human experts with backgrounds in psychology to rate these samples based on the same guideline provided to GPT-4. We conducted Chi-Square tests to compare the ratings provided by GPT-4 and those given by the human experts.

## Results

**Evaluation of EmoScan Screening Agent**

Table. 1 summarizes the screening performance comparison between EmoScan and the baselines based on weighted F1 (Schultebraucks et al., 2022). As demonstrated, EmoScan significantly outperformed all baselines with zero-shot, few-shot and chain of thought prompting (F1 = 0.7467), with improvements in the classification of both depressive (F1 = 0.6333) and anxiety disorders (F1 = 0.8567). This result may be attributed to EmoScan's more cautious approach in identifying cases as positive (i.e., with emotional disorders) compared to the baselines as the precision was much higher for both depressive and anxiety disorders. Meanwhile, although the introduction of fine-grained categories may have increased the difficulty of the second classification task, EmoScan's performance (F1 = 0.2567) still exhibited considerable improvements compared to the base model (the highest F1 = 0.0467), showing the efficacy of PsyInterview.

| Model | F1 weighted | Dep F1 | Anx F1 | Dep Recall | Dep Precision | Anx Recall | Anx Precision | Fine-grained F1-weighted |
|---|---|---|---|---|---|---|---|---|
| **EmoScan** | **.7467** | **.6333** | **.8567** | .7700 | **.5400** | .8467 | **.8667** | .2567 |
| Mistral-7B (zero-shot) | .2100** | .2100*** | .2100*** | .8700 | .1200 | .8467 | .1167 | .0467 |
| Mistral-7B (few-shot) | .1267*** | .1533*** | .1067*** | .2033 | .1233 | .1800 | .0767 | .0400 |
| Mistral-7B (CoT) | .2833*** | .2867** | .2867** | .5133 | .1933 | .5933 | .1900 | .1000 |
| Mistral-7B (few-shot + CoT) | .1267*** | .1300** | .1167*** | .1300 | .0900 | .2300 | .1067 | .0233 |
| GPT-4 (zero-shot) | .3800** | .3600*** | .4067** | .9233 | .2233 | .8967 | .2633 | .2667 |
| GPT-4 (few-shot) | .5900** | .6167 | .5600** | .9733 | .4500 | .8733 | .4133 | .3100 |
| GPT-4 (CoT) | .4467** | .4600* | .4300** | .9733 | .3033 | .8200 | .2933 | **.3700** |
| GPT-4 (few-shot + CoT) | .5133** | .4967* | .5367** | .9233 | .3400 | .8967 | .3800 | .3600 |
| Llama3 (zero-shot) | .3300** | .3033*** | .3633** | .9467 | .1767 | **.9733** | .2233 | .2600 |
| Llama3 (few-shot) | .4533** | .4100*** | .4933** | **1.0000** | .3100 | .8233 | .3500 | .3067 |
| Llama3 (CoT) | .4067** | .3533*** | .4600** | .9467 | .2167 | .9233 | .3100 | .2733 |
| Llama3 (few-shot + CoT) | .4933** | .4733*** | .5167** | .8967 | .3200 | .8500 | .3700 | .2933 |

Table. 1 Screening results comparison

*Note:* Dep = Depressive Disorders, Anx = Anxiety Disorders, CoT = chain of thought. The second to fourth columns listed the results of statistical significance tests between EmoScan and baselines. Bold numbers indicate the highest performance in the respective category. * $p < 0.05$, ** $p < 0.01$, *** $p < 0.001$.

When screening emotional disorders, an explanation for the screening result will be provided by EmoScan, which can help psychiatrists understand the underlying logic of the screening outputs generated by the LLMs. To assess the effectiveness of these explanations, we utilized multiple metrics: ROUGE (Lin, 2004), BLEU (Papineni, 2002), and BERTScore (Zhang et al., 2019). The ROUGE assesses the overlap between the generated and reference texts (i.e., truth explanations); BLEU evaluates textual similarity based on specific length's overlap; while BERTScore, known for capturing semantic similarity more effectively, focuses on contextual meaning. EmoScan demonstrated exceptional performance in BERTScore (0.9408), indicating a high level of semantic alignment. It also achieved high scores in BLEU (0.0660) for matching short sequences and ROUGE-1 (0.3951) for matching single words. These results underscore EmoScan's capability to produce accurate and contextually appropriate explanations. Although EmoScan's performance in ROUGE-2 (0.1132) and ROUGE-L (0.2086) metrics was slightly below that of Llama3, it outperformed all other baselines in the two metrics, showing its proficiency in capturing longer text dependencies. Overall, EmoScan's performance and explainability remain superior, as demonstrated in Table 2.

| Model | BERTScore | BLEU 2-gram | ROUGE-1 | ROUGE-2 | ROUGE-L |
| --- | --- | --- | --- | --- | --- |
| **EmoScan** | **0.9408** | **0.0660** | **0.3951** | 0.1132 | 0.2086 |
| Mistral-7B (zero-shot) | 0.6897 | 0.0252 | 0.2204 | 0.0539 | 0.1214 |
| Mistral-7B (few-shot) | 0.6110 | 0.0103 | 0.1686 | 0.0341 | 0.1041 |
| Mistral-7B (CoT) | 0.7259 | 0.0227 | 0.2258 | 0.0535 | 0.1330 |
| Mistral-7B (few-shot + CoT) | 0.4471 | 0.0075 | 0.0964 | 0.0166 | 0.0591 |
| GPT-4 (zero-shot) | 0.8968 | 0.0391 | 0.3248 | 0.0752 | 0.1674 |
| GPT-4 (few-shot) | 0.9188 | 0.0364 | 0.3191 | 0.0762 | 0.1661 |
| GPT-4 (CoT) | 0.9259 | 0.0451 | 0.3301 | 0.0860 | 0.1857 |
| GPT-4 (few-shot + CoT) | 0.9268 | 0.0414 | 0.3276 | 0.0871 | 0.1696 |
| Llama3 (zero-shot) | 0.8775 | 0.0555 | 0.3737 | **0.1135** | 0.2072 |
| Llama3 (few-shot) | 0.9321 | 0.0571 | 0.3655 | 0.1079 | **0.2101** |
| Llama3 (CoT) | 0.9309 | 0.0608 | 0.3724 | 0.1085 | 0.2086 |
| Llama3 (fewshot + CoT) | 0.9218 | 0.0543 | 0.3484 | 0.1051 | 0.2090 |

**Table. 2 Screening explanations on cases with emotional disorders**
*Note*: CoT = chain of thought

**Generalizability of EmoScan Screening Agent**

We compared the performance on the D4 dataset before and after training the model to highlight EmoScan's generalizability in related tasks. In the original study, psychiatrists categorized D4 cases into two groups: one with a risk of depression and the other without a depression risk. EmoScan (F1 = 0.67) outperformed the base model Mistral-7B (F1 = 0.64) when classifying the two groups.

**Evaluation of EmoScan Interviewing Agent**

EmoScan outperformed Mistral, Llama3, and GPT-4 in most dimensions of interviewing performance, as evident from both GPT-4 and human experts' rating results (Fig. 4). We allowed each rater to vote for one of the two conversations generated by either EmoScan or one of the baselines, and the decision could also result in a tie. As seen in Fig. 4, EmoScan had a higher winning rate than the baselines in dimensions such as medical history, family history of mental disorders, personal and social history, and warning that the interview is over, reflecting the interviewing efficacy of EmoScan. Moreover, the Chi-Square test revealed a significant correlation between the ratings by GPT-4 and human experts across all six dimensions: history of present illness ($X^2$ = 13.9763, $p$ = *.0074*), medical history ($X^2$ = 42.2244, $p$ < *.001*), personal and social history ($X^2$ = 9.6004, $p$ = *.0477*), family history of mental disorder ($X^2$ = 69.8743, $p$ < *.001*), warns that interview is over ($X^2$ = 40.3137, $p$ < *.001*), and conclusion with interest and appreciation ($X^2$ = 17.4907, $p$ = *.0016*). These results demonstrated the reliability of using GPT-4 instead of human experts for assessing clinical interviewing skills.

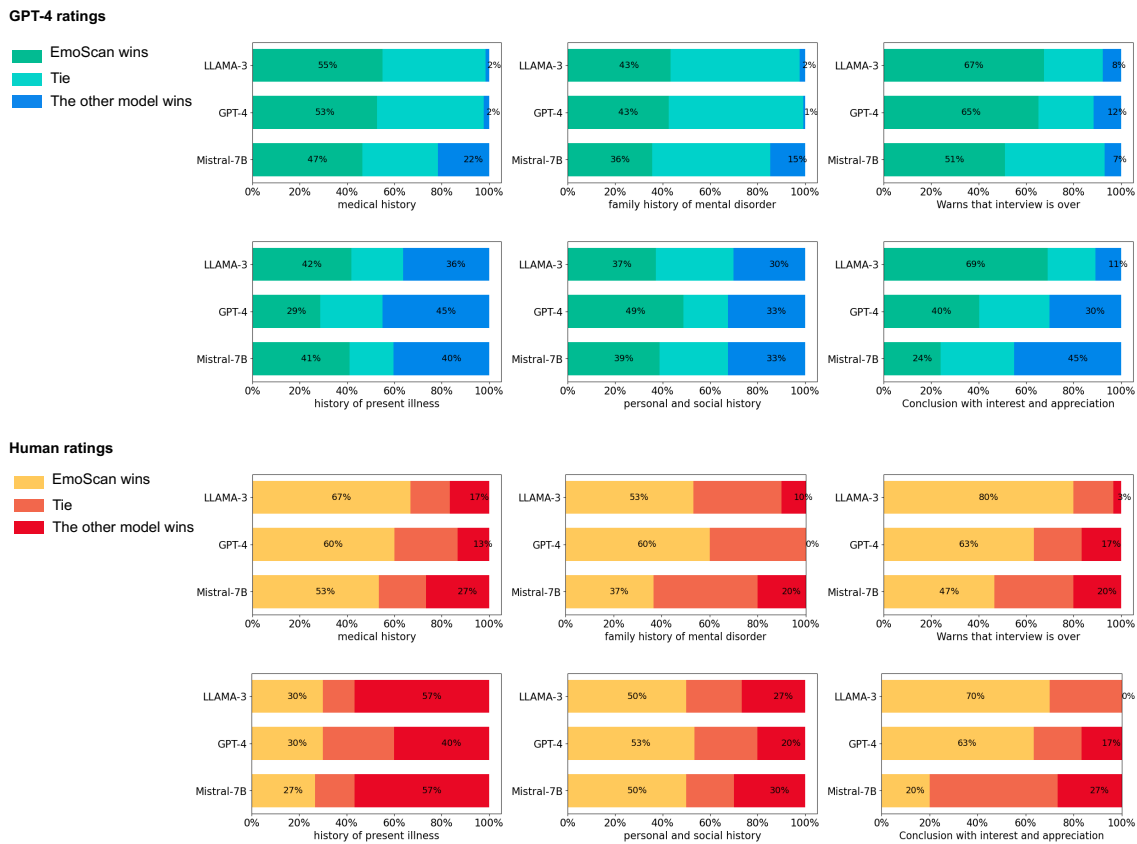

**Fig. 4 Ratings on Interviewing performance by GPT-4 and Human experts.** The blue-green section (above) shows the ratings by GPT-4, and the yellow-red section (below) displays the ratings provided by human experts. The correlation is significant across all six dimensions, with the highest p-value being less than 0.05.

## Discussion

The present study introduced a new pipeline for synthesizing clinical interviews to screen for emotional disorders, generated an interviewing dataset PsyInterview and trained a system, EmoScan, capable of both providing screening results with explanations and conducting interviews. Upon evaluation, EmoScan demonstrated superior performance in screening emotional disorders, providing robust screening explanations compared to baselines, conducting interviews, and showed greater generalizability than the base model. In a clinical context, EmoScan has the potential to save expert clinicians' time by effectively communicating with individuals and identifying those with emotional disorders. Additionally, it could also provide explanations for its output, enabling experts to comprehend the decision-making logic and gain confidence in the results.

Researchers had underscored the potential of AI to complement professional mental health expertise (Elyoseph et al., 2024), while emphasizing domain-specific scope to avoid risks associated with applications of general LLMs to clinical tasks (Au Yeung et al., 2023). Our findings highly supported the recent perspectives on utilizing LLMs for mental health applications. Our results showed that EmoScan achieved significantly higher overall F1 scores compared to all the baseline general LLMs, though its recall score was marginally lower than a few of them. This could be attributed to that EmoScan was a cautious system with high precision—especially when screening for anxiety disorders, where general LLMs tended to underperform and incorrectly label many healthy individuals as patients. As a cautious system, EmoScan minimized the likelihood of false diagnoses for emotional disorders, thus preventing unnecessary costs related to further treatment. In summary, EmoScan holds potential for application in clinical screening, offering valuable assistance to clinicians in making more efficient diagnoses.

One of our contributions was the development of a scalable pipeline for generating data efficiently. Previous research in this field had often relied on data that either incurred high costs or utilized pre-existing datasets annotated using questionnaires. For instance, the depression-related conversation corpus D4 (Yao et al., 2022) employed as a validation dataset in our study was created by recruiting individuals to role-play patients or doctors. While the data quality of D4 was good, the associated costs of human role-playing were relatively high. Other notable studies explored emotional disorders using conversational datasets such as DAIC-WOZ, a multi-modal dataset that was developed for depression diagnosis (Chen et al., 2024b; Gratch et al., 2014). Due to collection difficulties and privacy considerations, the sample size of such datasets was relatively small, limiting their application in automated clinical screening. In contrast, our pipeline made use of automated processes to convert large-scale clinical materials related to emotional disorders to usable training data. Our balanced approach, incorporating both automation and selective human monitoring, not only ensured high data quality but also significantly reduced costs and ethical concerns associated with real-life conversations. Consequently, our pipeline provided researchers and clinical experts with a tool for training specialized LLMs using their clinical data, thereby bypassing the high financial burden and ethical limitations tied to the use of real-life conversations.

One limitation of our study was the small sample size of each fine-grained emotional disorder. While our dataset comprised around 20 different emotional disorders, the number of samples for each fine-grained disorder was limited due to the constraints of available sources. The limited training sample size for each disorder posed a challenge for many LLMs, including EmoScan, as it prevented them from learning hidden patterns, which ultimately resulted in lower accuracy when identifying each fine-grained disorder. To improve model performance, future studies could consider enlarging the sample size for each fine-grained disorder. Additionally, future researchers may integrate multi-modality information to train LLMs. For example, some previous research had underscored the benefits of incorporating acoustic speech information within LLM frameworks for depression detection (Chen et al., 2024b; Zhang et al., 2024; Tao et al., 2023). Our study focused on textual data for easier implementation, yet incorporating multi-modal inputs could help researchers enhance screening accuracy.

## Conclusions

Our study introduced a novel pipeline for effective screening and interviewing of emotional disorders using Large Language Models (LLMs). Using our pipeline, we synthesized data from existing clinical interviews and created the PsyInterview dataset. Subsequently, we developed EmoScan, an LLM-based system trained on our collected dataset for screening emotional disorders with explanations and conducting interviews to collect clinical information. EmoScan demonstrated superior accuracy, robust explanations, and strong interviewing skills, significantly outperforming baseline models. Our pipeline and models hold considerable potential for efficient and effective clinical screening and interviewing, providing mental health professionals with a valuable tool to support their work.

# References

Adarmouch, L., Felaefel, M., Wachbroit, R., & Silverman, H. (2020). Perspectives regarding privacy in clinical research among research professionals from the Arab region: an exploratory qualitative study. *BMC medical ethics*, *21*, 1-16.

American Psychiatric Association. (2015). *Structured clinical interview for DSM-5 (SCID-5)*. Retrieved from http://www.appi.org/products/structured-clinical-interview-for-dsm-5-scid-5

Au Yeung, J., Kraljevic, Z., Luintel, A., Balston, A., Idowu, E., Dobson, R. J., & Teo, J. T. (2023). AI chatbots not yet ready for clinical use. *Frontiers in digital health*, *5*, 1161098.

Blank, I. A. (2023). What are large language models supposed to model?. *Trends in Cognitive Sciences*.

Brown, T., Mann, B., Ryder, N., Subbiah, M., Kaplan, J. D., Dhariwal, P., ... & Amodei, D. (2020). Language models are few-shot learners. *Advances in neural information processing systems*, *33*, 1877-1901.

Bullis, J. R., Boettcher, H., Sauer-Zavala, S., Farchione, T. J., & Barlow, D. H. (2019). What is an emotional disorder? A transdiagnostic mechanistic definition with implications for assessment, treatment, and prevention. *Clinical psychology: Science and practice*, *26*(2), e12278.

Chen, S., Wang, M., Lv, M., Zhang, Z., Juqianqian, J., Dejiyangla, D., ... & Wu, M. (2024a, June). Mapping Long-term Causalities in Psychiatric Symptomatology and Life Events from Social Media. In *Proceedings of the 2024 Conference of the North American Chapter of the Association for Computational Linguistics: Human Language Technologies (Volume 1: Long Papers)* (pp. 5472-5487).

Chen, Z., Deng, J., Zhou, J., Wu, J., Qian, T., & Huang, M. (2024b, June). Depression detection in clinical interviews with LLM-empowered structural element graph. In *Proceedings of the 2024 Conference of the North American Chapter of the Association for Computational Linguistics: Human Language Technologies (Volume 1: Long Papers)* (pp. 8174-8187).

Cheng, J., Sabour, S., Sun, H., Chen, Z., & Huang, M. (2022). Pal: Persona-augmented emotional support conversation generation. *arXiv preprint arXiv:2212.09235*.

Cong, Y., LaCroix, A. N., & Lee, J. (2024). Clinical efficacy of pre-trained large language models through the lens of aphasia. *Scientific Reports*, *14*(1), 15573.

Elyoseph, Z., Levkovich, I., & Shinan-Altman, S. (2024). Assessing prognosis in depression: comparing perspectives of AI models, mental health professionals and the general public. *Family Medicine and Community Health*, *12*(Suppl 1).

Farruque, N., Goebel, R., Sivapalan, S., & Zaïane, O. R. (2024). Depression symptoms modelling from social media text: an LLM driven semi-supervised learning approach. *Language Resources and Evaluation*, 1-29.

Finning, K., Moore, D., Ukoumunne, O. C., Danielsson-Waters, E., & Ford, T. (2017). The association between child and adolescent emotional disorder and poor attendance at school: a systematic review protocol. *Systematic reviews*, *6*, 1-5.

Gratch, J., Artstein, R., Lucas, G. M., Stratou, G., Scherer, S., Nazarian, A., ... & Morency, L. P. (2014, May). The distress analysis interview corpus of human and computer interviews. In *LREC* (pp. 3123-3128).

Hackl, V., Müller, A. E., Granitzer, M., & Sailer, M. (2023). Is GPT-4 a reliable rater? Evaluating consistency in GPT-4's text ratings. In *Frontiers in Education* (Vol. 8, p. 1272229). Frontiers Media SA.

Institute for Health Metrics and Evaluation. (2019). *GBD results*. https://vizhub.healthdata.org/gbd-results/

Jansson, L., & Nordgaard, J. (2016). *The psychiatric interview for differential diagnosis* (Vol. 270). Switzerland: Springer.

Jiang, A. Q., Sablayrolles, A., Mensch, A., Bamford, C., Chaplot, D. S., Casas, D. D. L., ... & Sayed, W. E. (2023). Mistral 7B. *arXiv preprint arXiv:2310.06825*.

Ke, L., Tong, S., Chen, P., & Peng, K. (2024). Exploring the frontiers of llms in psychological applications: A comprehensive review. *arXiv preprint arXiv:2401.01519*.

King, D. R., Nanda, G., Stoddard, J., Dempsey, A., Hergert, S., Shore, J. H., & Torous, J. (2023). An introduction to generative artificial intelligence in mental health care: considerations and guidance. *Current psychiatry reports*, *25*(12), 839-846.

Konnopka, A., & König, H. (2020). Economic burden of anxiety disorders: a systematic review and meta-analysis. *Pharmacoeconomics*, *38*, 25-37.

Lan, X., Cheng, Y., Sheng, L., Gao, C., & Li, Y. (2024). Depression Detection on Social Media with Large Language Models. *arXiv preprint arXiv:2403.10750*.

Liao, Y., Meng, Y., Wang, Y., Liu, H., Wang, Y., & Wang, Y. (2024). Automatic Interactive Evaluation for Large Language Models with State Aware Patient Simulator. *arXiv preprint arXiv:2403.08495*.

Lin, C. Y. (2004). Rouge: A package for automatic evaluation of summaries. In *Text summarization branches out* (pp. 74-81).

Liu, S., Zheng, C., Demasi, O., Sabour, S., Li, Y., Yu, Z., ... & Huang, M. (2021). Towards emotional support dialog systems. *arXiv preprint arXiv:2106.01144*.

Longwell, J. B., Hirsch, I., Binder, F., Conchas, G. A. G., Mau, D., Jang, R., ... & Grant, R. C. (2024). Performance of Large Language Models on Medical Oncology Examination Questions. *JAMA Network Open*, *7*(6), e2417641-e2417641.

Morrison, J. (2016). *The first interview*. Guilford Publications.

Morrison, J. (2023). *Diagnosis made easier: Principles and techniques for mental health clinicians*. Guilford Publications.

Omar Sr, M., Soffer Sr, S., Charney Sr, A., Landi, I., Nadkarni, G., & Klang Jr, E. (2024). Applications of Large Language Models in Psychiatry: A Systematic Review. *medRxiv*, 2024-03.

OpenAI. 2023. Gpt-4 technical report.

Papineni, K., Roukos, S., Ward, T., & Zhu, W. J. (2002). Bleu: a method for automatic evaluation of machine translation. In *Proceedings of the 40th annual meeting of the Association for Computational Linguistics* (pp. 311-318).

Prendergast, K. (2018). *Psychiatric case studies for advanced practice*. Lippincott Williams & Wilkins.

Rosenman, G., Wolf, L., & Hendler, T. (2024). LLM Questionnaire Completion for Automatic Psychiatric Assessment. *arXiv preprint arXiv:2406.06636*.

Santomauro, D. F., Herrera, A. M. M., Shadid, J., Zheng, P., Ashbaugh, C., Pigott, D. M., ... & Ferrari, A. J. (2021). Global prevalence and burden of depressive and anxiety disorders in 204 countries and territories in 2020 due to the COVID-19 pandemic. *The Lancet*, *398*(10312), 1700-1712.

Sau, A., & Bhakta, I. (2019). Screening of anxiety and depression among seafarers using machine learning technology. *Informatics in Medicine Unlocked*, *16*, 100228.

Schultebraucks, K., Yadav, V., Shalev, A. Y., Bonanno, G. A., & Galatzer-Levy, I. R. (2022). Deep learning-based classification of posttraumatic stress disorder and depression following trauma utilizing visual and auditory markers of arousal and mood. *Psychological Medicine*, *52*(5), 957-967.

Stade, E. C., Stirman, S. W., Ungar, L. H., Boland, C. L., Schwartz, H. A., Yaden, D. B., ... & Eichstaedt, J. C. (2024). Large language models could change the future of behavioral healthcare: a proposal for responsible development and evaluation. *NPJ Mental Health Research*, *3*(1), 12.

Tao, Y., Yang, M., Shen, H., Yang, Z., Weng, Z., & Hu, B. (2023, December). Classifying anxiety and depression through LLMs virtual interactions: A case study with ChatGPT. In *2023 IEEE International Conference on Bioinformatics and Biomedicine (BIBM)* (pp. 2259-2264). IEEE.

Tu, T., Palepu, A., Schaekermann, M., Saab, K., Freyberg, J., Tanno, R., ... & Natarajan, V. (2024). Towards conversational diagnostic ai. *arXiv preprint arXiv:2401.05654*

Wang, J., Yao, Z., Yang, Z., Zhou, H., Li, R., Wang, X., ... & Yu, H. (2023). Notechat: A dataset of synthetic doctor-patient conversations conditioned on clinical notes. *arXiv preprint arXiv:2310.15959*.

Wang, Y., Inkpen, D., & Gamaarachchige, P. K. (2024, March). Explainable depression detection using large language models on social media data. In *Proceedings of the 9th Workshop on Computational Linguistics and Clinical Psychology (CLPsych 2024)* (pp. 108-126).

Watson, D., O'Hara, M. W., & Stuart, S. (2008). Hierarchical structures of affect and psychopathology and their implications for the classification of emotional disorders. *Depression and Anxiety*, *25*(4), 282-288.

Wei, J., Wang, X., Schuurmans, D., Bosma, M., Xia, F., Chi, E., ... & Zhou, D. (2022). Chain-of-thought prompting elicits reasoning in large language models. *Advances in neural information processing systems*, *35*, 24824-24837.

Wright, B., Dave, S., & Dogra, N. (2017). *100 cases in psychiatry*. CRC Press.

Wu, Q., Bansal, G., Zhang, J., Wu, Y., Zhang, S., Zhu, E., ... & Wang, C. (2023). Autogen: Enabling next-gen llm applications via multi-agent conversation framework. *arXiv preprint arXiv:2308.08155*.

Yao, B., Shi, C., Zou, L., Dai, L., Wu, M., Chen, L., ... & Yu, K. (2022). D4: a chinese dialogue dataset for depression-diagnosis-oriented chat. *arXiv preprint arXiv:2205.11764*.

Xu, X., Yao, B., Dong, Y., Gabriel, S., Yu, H., Hendler, J., ... & Wang, D. (2024). Mental-llm: Leveraging large language models for mental health prediction via online text data. *Proceedings of the ACM on Interactive, Mobile, Wearable and Ubiquitous Technologies*, *8*(1), 1-32.

Zhang, T., Kishore, V., Wu, F., Weinberger, K. Q., & Artzi, Y. (2019). Bertscore: Evaluating text generation with bert. *arXiv preprint arXiv:1904.09675*.

Zhang, X., Liu, H., Xu, K., Zhang, Q., Liu, D., Ahmed, B., & Epps, J. (2024). When LLMs meets acoustic landmarks: An efficient approach to integrate speech into large language models for depression detection. *arXiv preprint arXiv:2402.13276*.

Zvolensky, M. J., Farris, S. G., Leventhal, A. M., & Schmidt, N. B. (2014). Anxiety sensitivity mediates relations between emotional disorders and smoking. *Psychology of Addictive Behaviors*, *28*(3), 912.